\documentclass[letterpaper, 10 pt, conference]{ieeeconf} 
\IEEEoverridecommandlockouts
\usepackage{amsfonts}
\usepackage{graphics}
\usepackage{graphicx}
\usepackage{cite}
\usepackage{subfigure}
\usepackage{xcolor}
\usepackage{ulem}
\usepackage{algpseudocode}
\usepackage{algorithm}

\usepackage[font=footnotesize]{caption}
\def\baselinestretch{0.97}

\title{\LARGE \bf Ergodic imitation: Learning from what to do and what \textit{not} to do}

\author{Aleksandra Kalinowska*$^{1}$, Ahalya Prabhakar*$^{1}$\thanks{*Authors contributed equally}, Kathleen Fitzsimons$^{1}$, and Todd Murphey$^{1,2}$%
	\thanks{This material is based upon work supported by the NSF under Grant CNS 1837515. Any opinions, findings and conclusions or recommendations expressed in this material are those of the authors and do not necessarily reflect the views of the aforementioned institutions.}
    \thanks{$^{1}$Mechanical Engineering, Northwestern University,
		Evanston, IL}
	\thanks{$^{2}$Physical Therapy and Human Movement Sciences, Northwestern University, Chicago, IL}
}

\begin{document}
\maketitle
\thispagestyle{empty}
\pagestyle{empty}

%===============================================================================

\begin{abstract}
With growing access to versatile robotics, it is beneficial for end users to be able to teach robots tasks without needing to code a control policy. One possibility is to teach the robot through successful task executions. However, near-optimal demonstrations of a task can be difficult to provide and even successful demonstrations can fail to capture task aspects key to robust skill replication. Here, we propose a learning from demonstration (LfD) approach that enables learning of robust task definitions without the need for near-optimal demonstrations. We present a novel algorithmic framework for learning tasks based on the ergodic metric---a measure of information content in motion. Moreover, we make use of negative demonstrations---demonstrations of what \textit{not} to do---and show that they can help compensate for imperfect demonstrations, reduce the number of demonstrations needed, and highlight crucial task elements improving robot performance. In a proof-of-concept example of cart-pole inversion, we show that negative demonstrations alone can be sufficient to successfully learn and recreate a skill. Through a human subject study with 24 participants, we show that consistently more information about a task can be captured from combined positive and negative (\textit{posneg}) demonstrations than from the same amount of just positive demonstrations. Finally, we demonstrate our learning approach on simulated tasks of target reaching and table cleaning with a 7-DoF Franka arm. Our results point towards a future with robust, data-efficient LfD for novice users.

\end{abstract}

% Two or three meaningful keywords should be added here
% \keywords{Imitation Learning, Data-efficient LfD, Assistive Robotics} 

%=================================================================

\section{Introduction}
	
Many assistive robots being deployed in people’s homes or on factory floors are capable of performing a variety of tasks. As such, it is beneficial for end users to be able to customize these robots by teaching them tasks specific to their needs. However, it is often not possible to provide high-quality task demonstrations. This could be because the task is challenging for a person to perform, e.g., cart-pole inversion due to its unintuitive dynamics, or the person is limited by a low-dimensional control interface, such as a joystick, for providing demonstrations to a 7-DoF robotic arm. Although successful approaches exist for imitation learning, including Dynamic Motion Primitives (DMPs)~\cite{ijspeert2013dynamical}, inverse reinforcement learning (IRL)~\cite{abbeel2004apprenticeship}, and others---as we describe in more detail in Section~\ref{sec:relevantwork}---few of the LfD frameworks allow for reliable learning from novice task demonstrations. 

Our approach stems from the idea of evaluating how much information about a task is encoded in motion---quantifying this using a measure of ergodicity. We propose ergodic imitation for robust learning from imperfect demonstrations. We define tasks through spatial distributions in state-based feature space. Through successive demonstrations, we learn the underlying distribution corresponding to a task and generate robot behavior via ergodic control~\cite{mavrommati2018real} with respect to the learned distributions. This learning framework allows us to combine multiple novice demonstrations into a successful objective and use model predictive control (MPC) to recreate trajectories for new, previously unencountered scenarios. It is worth noting that ergodic imitation does not focus on imitating trajectories directly---instead it emphasizes imitating trajectory \textit{statistics}. As a result, the method learns well from imperfect demonstrations and is robust to noise in individual demonstrations (e.g., corrective motions or perturbations). 

Moreover, we propose imitation learning using negative demonstrations. In some cases, it might be easier for a person to demonstrate what \textit{not} to do rather than to provide an exemplary task execution. Demonstrating things to avoid is something that people already intuitively do when teaching new skills to others. Robotic LfD can also largely benefit from incorporating negative demonstrations into the learning process. Ergodic imitation is a particularly suitable algorithmic framework, because it enables combining positive and negative demonstrations into a well-posed task objective. 

As part of this study, we validate our learning approach on two test beds: a virtual 2-D cart-pole system and a simulated 7-DoF robotic arm. We find that ergodic imitation (1) enables robust skill reconstruction that outperforms the provided demonstrations and (2) generalizes to different robot tasks. Additionally, we test the utility of negative demonstrations in an experiment with 24 participants. Our results show that there is consistent benefit to soliciting combined \textit{posneg} demonstrations compared to only positive demonstrations. 

 \begin{figure*}[t]
  \centering
    \includegraphics[width=\textwidth]{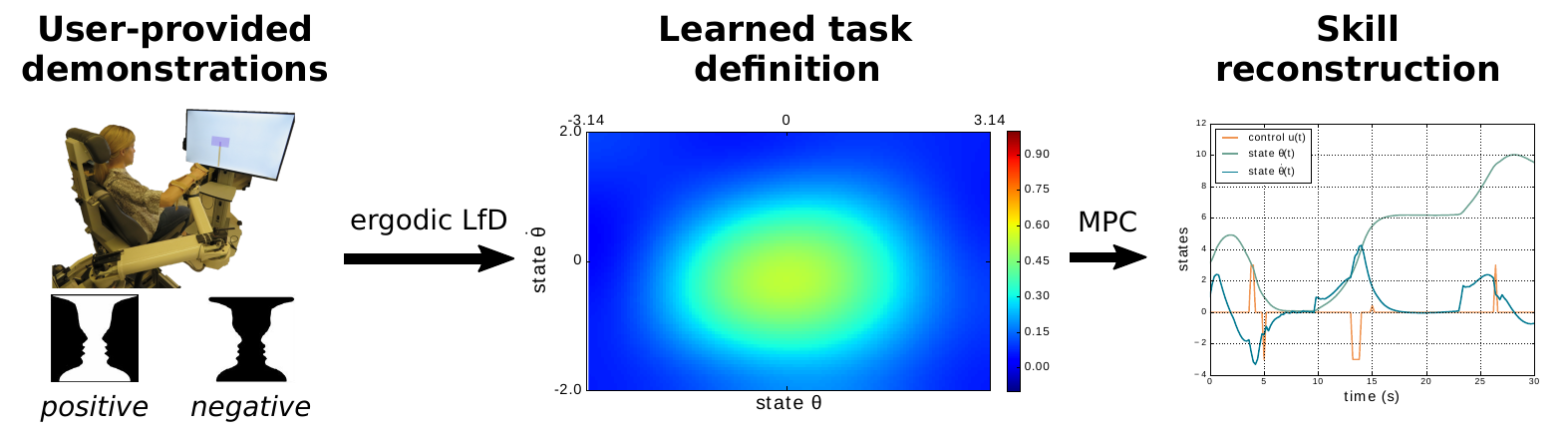}
    \caption{An overview of the learning process using ergodic imitation on the example of the cart-pole inversion task. Positive and negative demonstrations are combined to form \textit{posonly}, \textit{negonly}, or \textit{posneg} task definitions.}
     \label{fig: ergodic-LfD}
\end{figure*}

\section{Related Work}
\label{sec:relevantwork}

Most inverse reinforcement learning (IRL) and inverse optimal control (IOC) methods assume that the demonstrations representing the task are generated by an expert demonstrator providing optimal (or near-optimal) strategies for the task in order to generate feasible solutions~\cite{abbeel2004apprenticeship, levine2011nonlinear, NIPS2016_6391}. Other approaches---such as those that use probabilistic methods to learn a task~\cite{paraschos2013probabilistic,Kulak20RSS,calinon2016tutorial,schneider2010robot,jaquier2019learning}---also rely on highly skilled demonstrators, accounting for imperfections with relatively small-scale noise in the probabilistic representation. To enable task learning that more closely captures human preferences and accounts for imperfect or incomplete demonstrations, active learning methods have been developed~\cite{basu2019active, pmlr-v100-b-iy-ik20a, basu2018learning, RSS2020-LfD, mourad2020learning}. In these approaches, the human is treated as an oracle that the autonomy can query, improving learning quality. Alternatively in~\cite{ILconfidencescores}, the person provides ratings (or confidence scores) to a subset of demonstrations, quantifying the demonstrations' quality and increasing learning reliability. However, there is an inherent cost of time and effort to querying the user for input and so the user-in-the-loop learning process can be prohibitively frustrating~\cite{amershi2014power, cakmak2010designing}. The user tends to have a preference towards online learning approaches (e.g.,~\cite{RSS2020-LfD}) that do not require \textit{post-hoc} corrections to the learned robot policies. Therefore, in this work, we improve task learning during run-time by soliciting different types of demonstrations from the human teacher. 

Suboptimal and failed demonstrations have been used in LfD before and have been shown to improve the learning process~\cite{IRLfailure, TRO2017badexperiences}. In~\cite{IRLfailure}, the proposed approach learns faster and generalizes better than the original IRL method, because it is able to learn from failure. In~\cite{TRO2017badexperiences}, authors demonstrate that accommodating failed demonstrations improves learning particularly for multimodal tasks (a result we also observe in our experiments). In~\cite{AudeDonutConf,AudeDonutJournal}, researchers propose a method that can learn solely from failed demonstrations, acknowledging that combining successful and failed demonstrations would likely deliver highest quality results. Building on this prior work, we demonstrate the value of soliciting negative demonstrations---explicit demonstrations of what \textit{not} to do. We present an LfD algorithm that allows for combining positive (successful but possibly suboptimal) demonstrations to be combined with negative (unsuccessful and/or explicitly suboptimal) demonstrations together in the learning process.

While the above mentioned methods can successfully learn various skills, many cannot generate safety guarantees for the learned policy nor guarantee the dynamic feasibility of the generated trajectories~\cite{IRLfailure, TRO2017badexperiences, AudeDonutJournal}. In contrast, IOC methods, such as~\cite{IOC2020CoRL, IOC2017marctoussaint}, which use optimal control to generate actions, can have provable guarantees on feasibility and performance. Similarily, the proposed algorithm inherits formal properties from ergodic control and standard MPC methods. (1) The ergodic cost is globally convex w.r.t. distributions so long as the metric used is on a Sobolev space. Here, we use the spectral approach as in~\cite{mathew2011}. (2) Ergodic imitation inherits asymptotic convergence from ergodic control~\cite{mavrommati2018real}. In the cart-pole example, this implies that when the goal distribution is defined as a delta function at the unstable equilibrium, the statistics of the trajectory will asymptotically approach the delta function---the pole could occasionally fall, but the amount of time spent at the inverted equilibrium will approach 100\% as $t\to\infty$. (3) Safety sets can be specified through the use of barrier functions.

In this paper, we propose and validate ergodic imitation as one possible approach that allows for learning from both suboptimal and negative demonstrations by allowing demonstration-based distributions to be simply added together, and it maintains desired properties of existing IOC methods. We obtain these properties by defining objectives as state distributions and employing ergodic control for closed-loop trajectory generation. While other methods, such as IRL, could be adapted to learn an objective function over distributions, the nominal complexity of IRL is already exponential ($O(n^2 log (nk))$)~\cite{komanduru2019}. If one were to perform IRL over the set of distributions, the algorithm would further increase in computational complexity likely becoming intractable. In ergodic imitation, the learning step (in Eq.~(\ref{phik})) has linear time complexity, allowing us to avoid the computational complexity of IRL. As documented in prior work, ergodic control scales well to relatively high-dimensional spaces---it has been implemented on the half cheetah example with 26 observables and continuous actions~\cite{abraham2020KL}.

\begin{figure*}[t!]
\vspace{1em}
  \centering
    \begin{subfigure}%
    \centering
    \includegraphics[width=.425\textwidth]{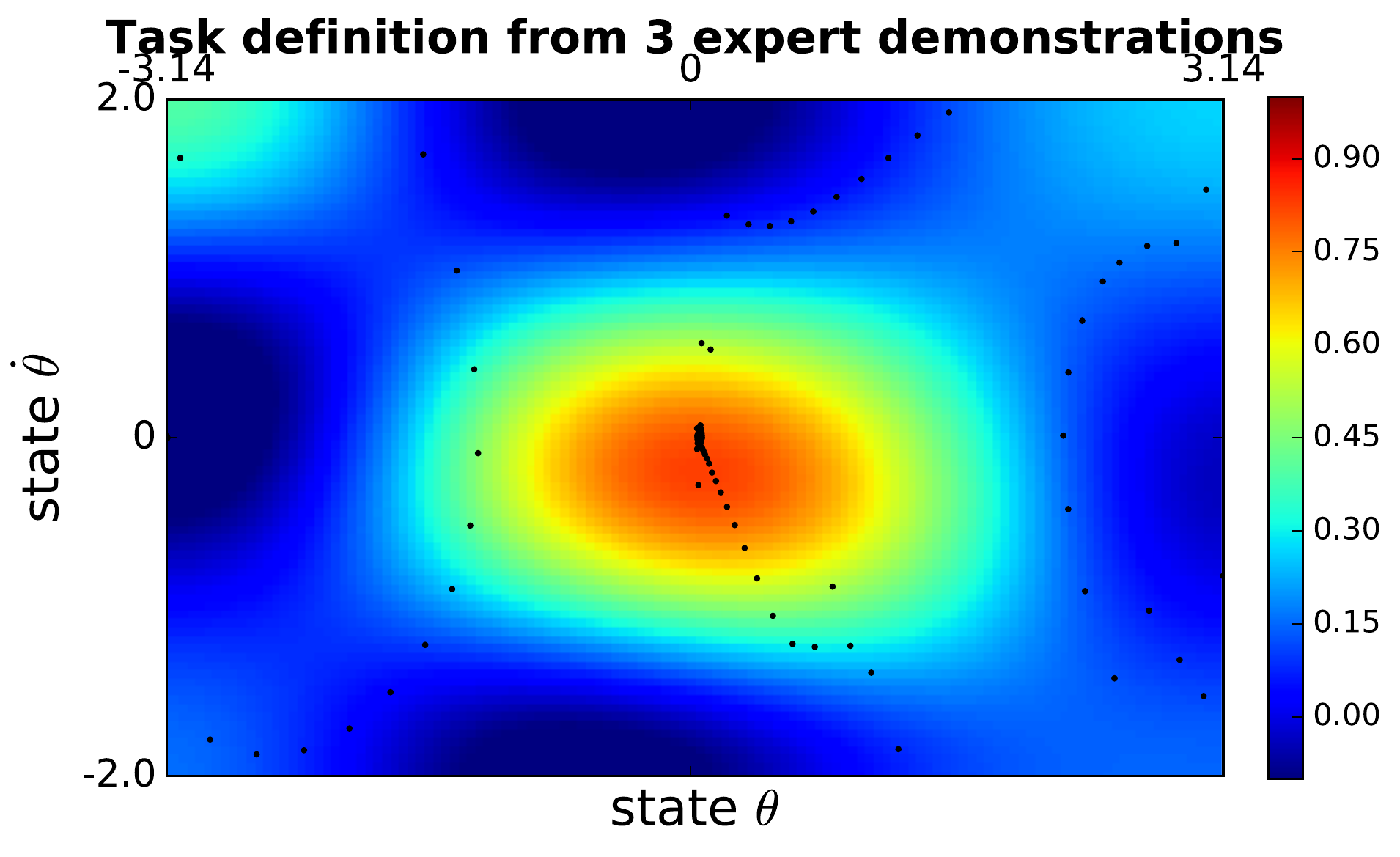}
  \end{subfigure}
  \begin{subfigure}%
    \centering
    \includegraphics[width=.32\textwidth]{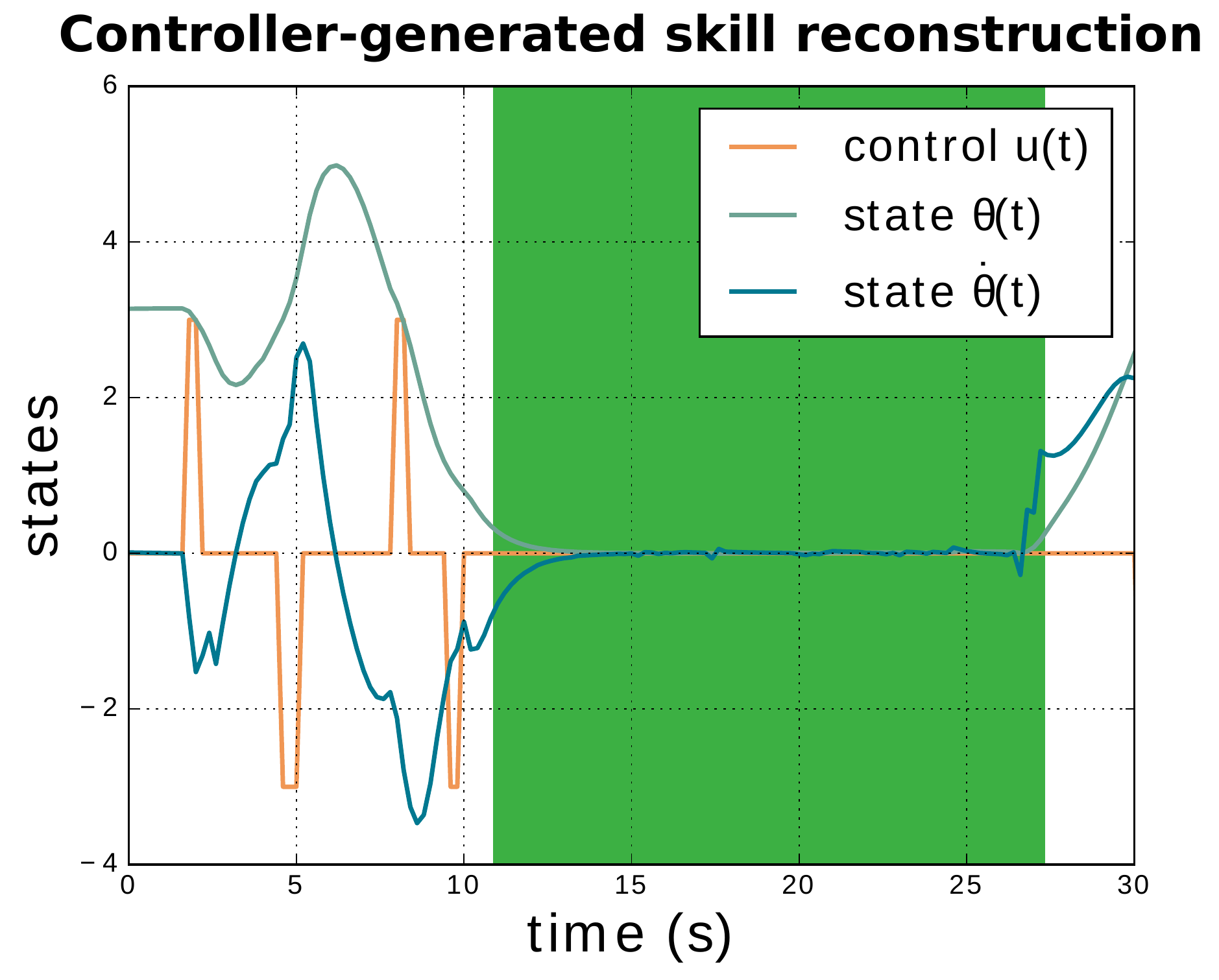}
  \end{subfigure}%
  \caption{Example task definition (left) and skill reconstruction (right) learned from 3 expert trajectories. An optimal controller is used to recreate the task given the learned goal distribution. Green indicates success---time when the cart-pole is inverted. The controller-generated trajectory is plotted on the right and overlayed on the task distribution on the left with black dots---note that the trajectory closely represents the underlying distribution subject to constraints imposed by system dynamics. }
  \label{fig: pos-expert}
\end{figure*}

\begin{figure*}[t]
  \centering
    \begin{subfigure} \centering
    \includegraphics[width=.3\textwidth]{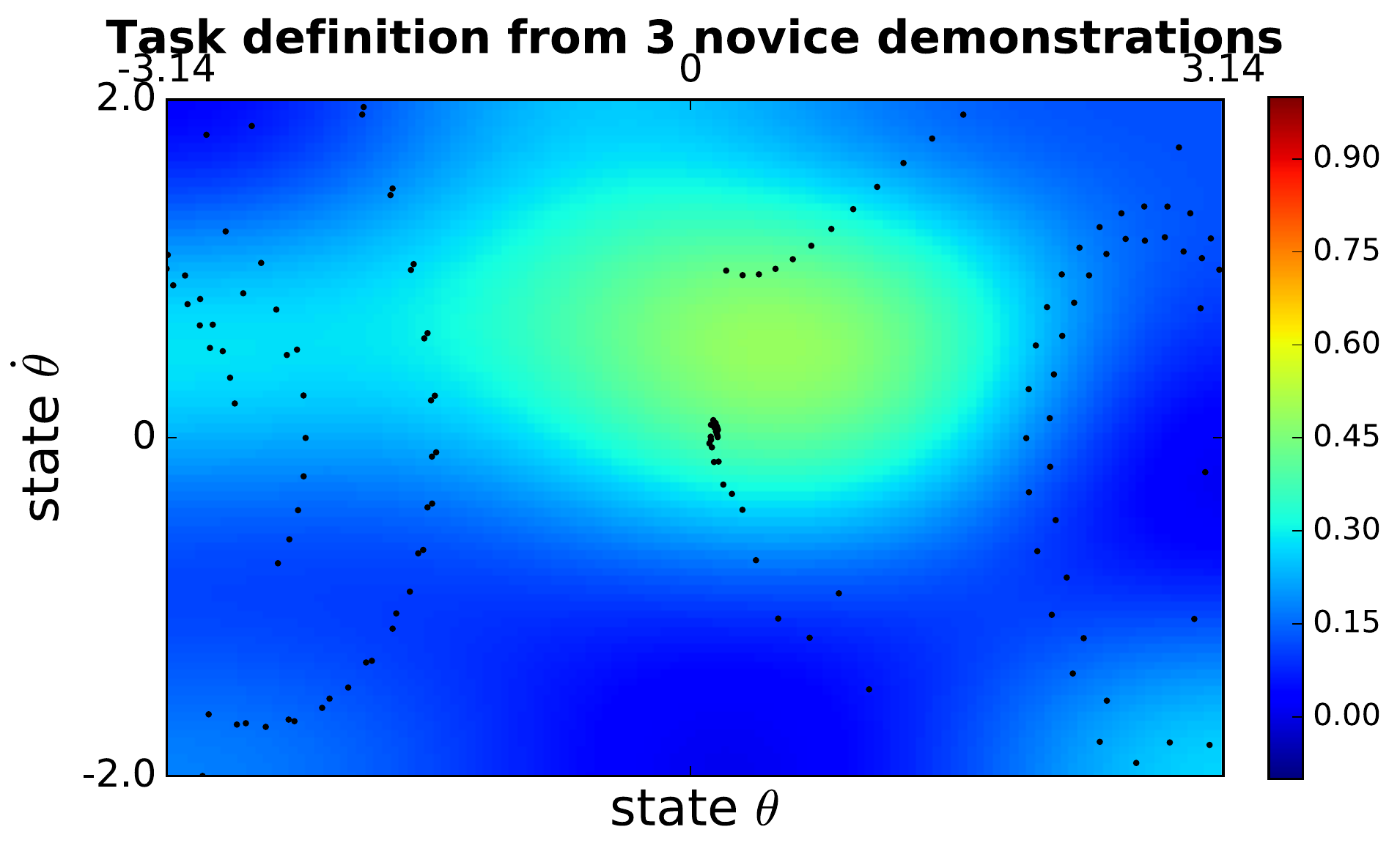}
    \end{subfigure} \hfill
    \begin{subfigure} \centering
    \includegraphics[width=.3\textwidth]{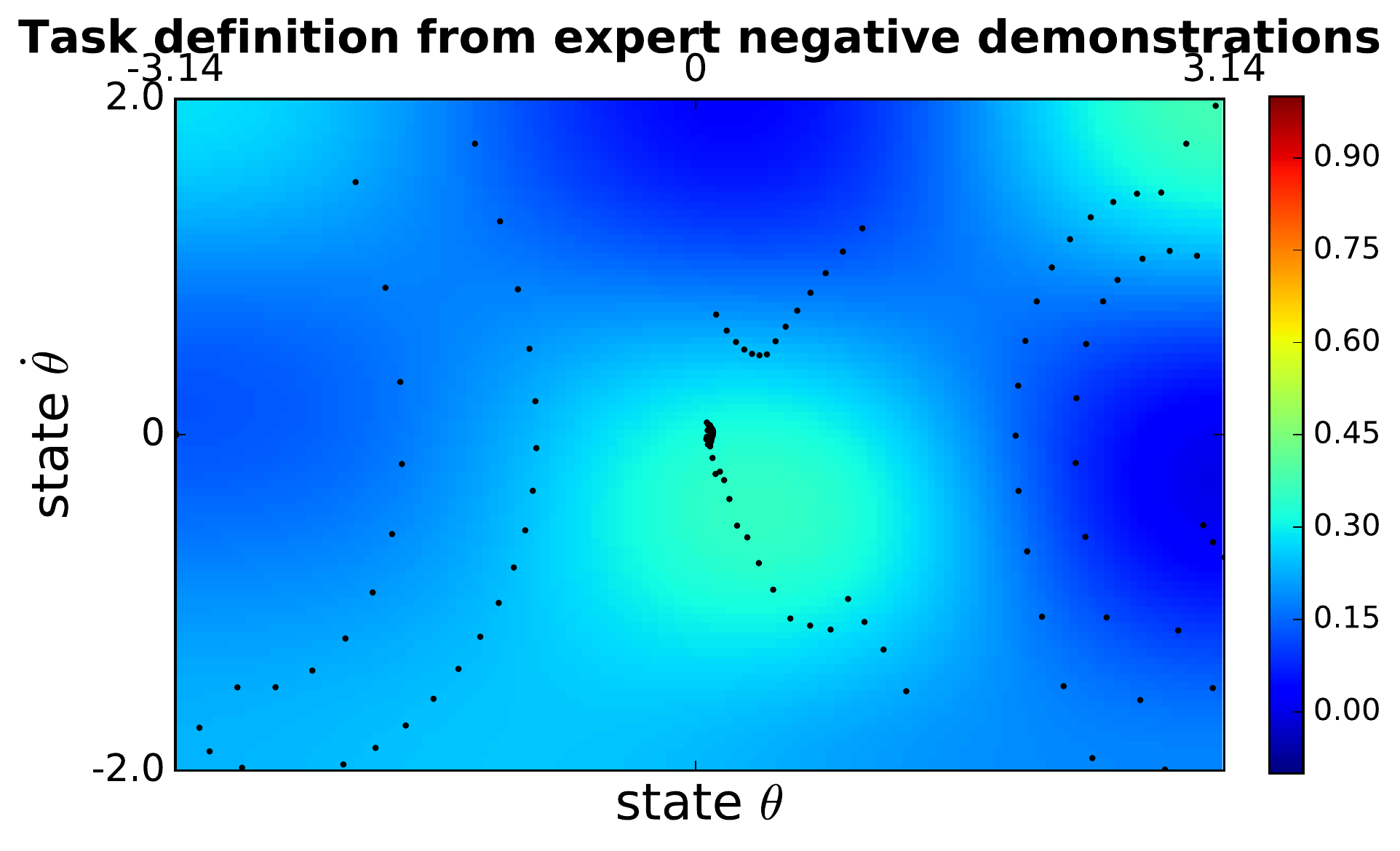} 
    \end{subfigure} \hfill
    \begin{subfigure} \centering
    \includegraphics[width=.3\textwidth]{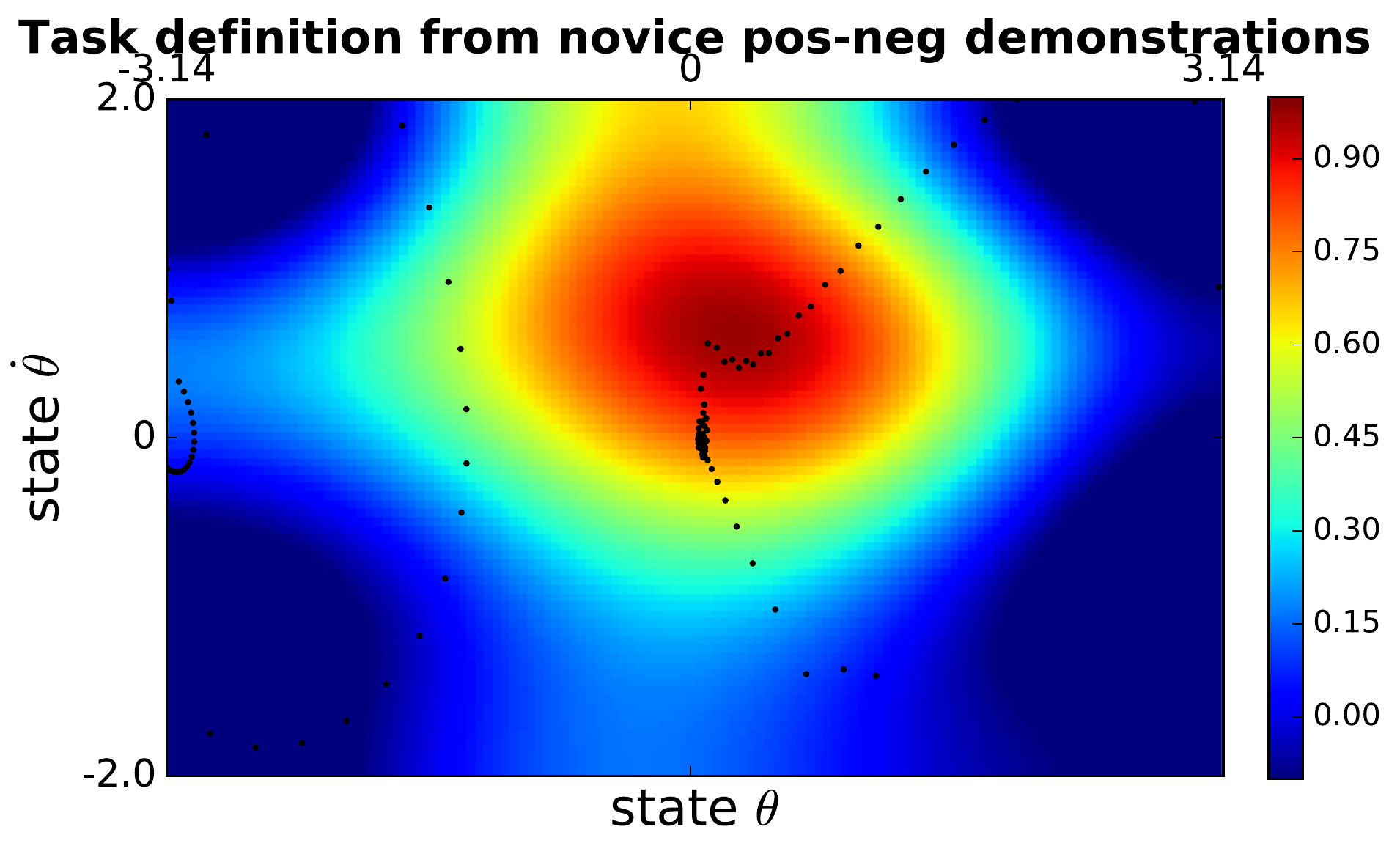}
    \end{subfigure}
    \\
    \begin{subfigure} \centering
    \includegraphics[width=0.3\textwidth]{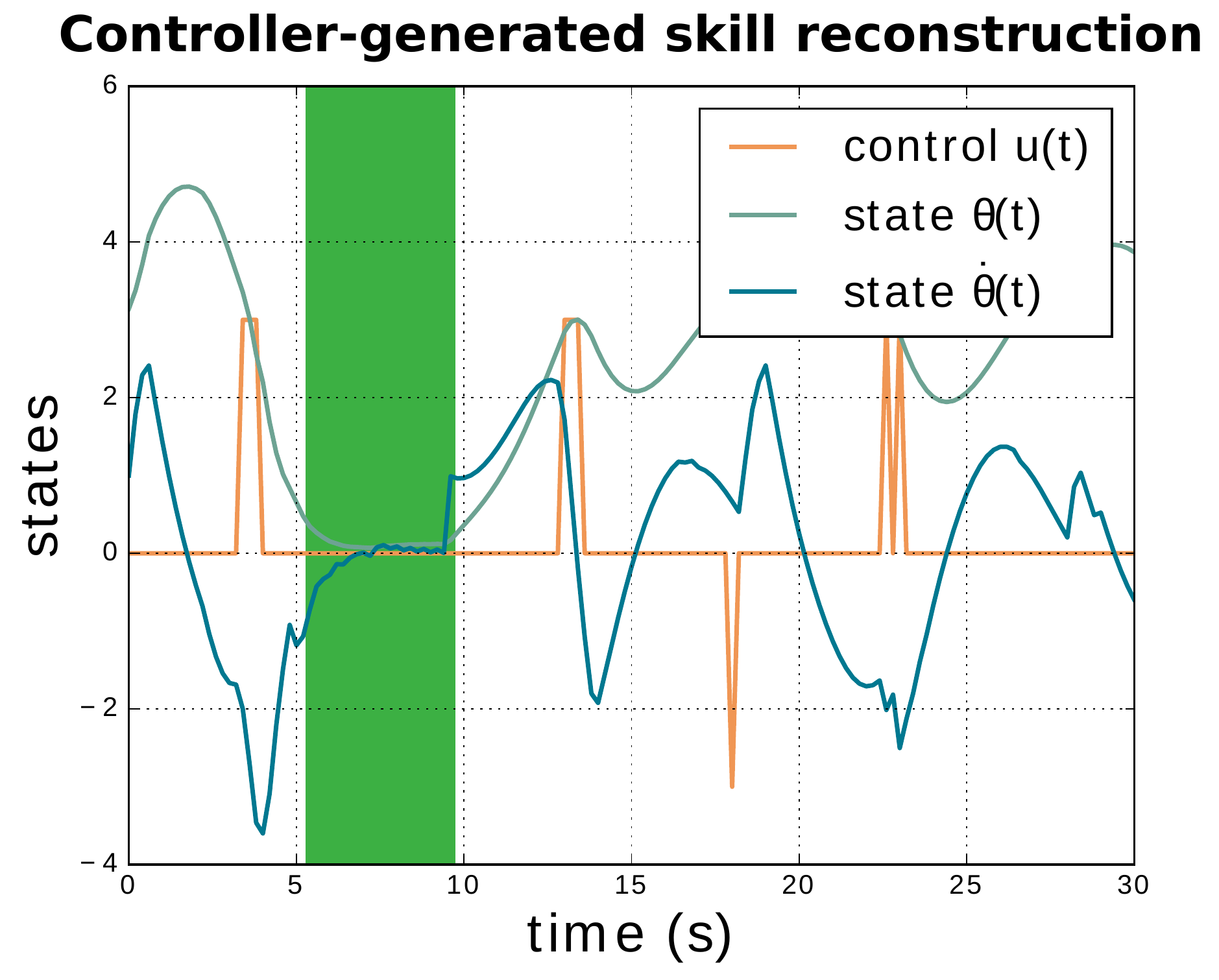} 
    \end{subfigure} \hfill
    \begin{subfigure} \centering
    \includegraphics[width=0.3\textwidth]{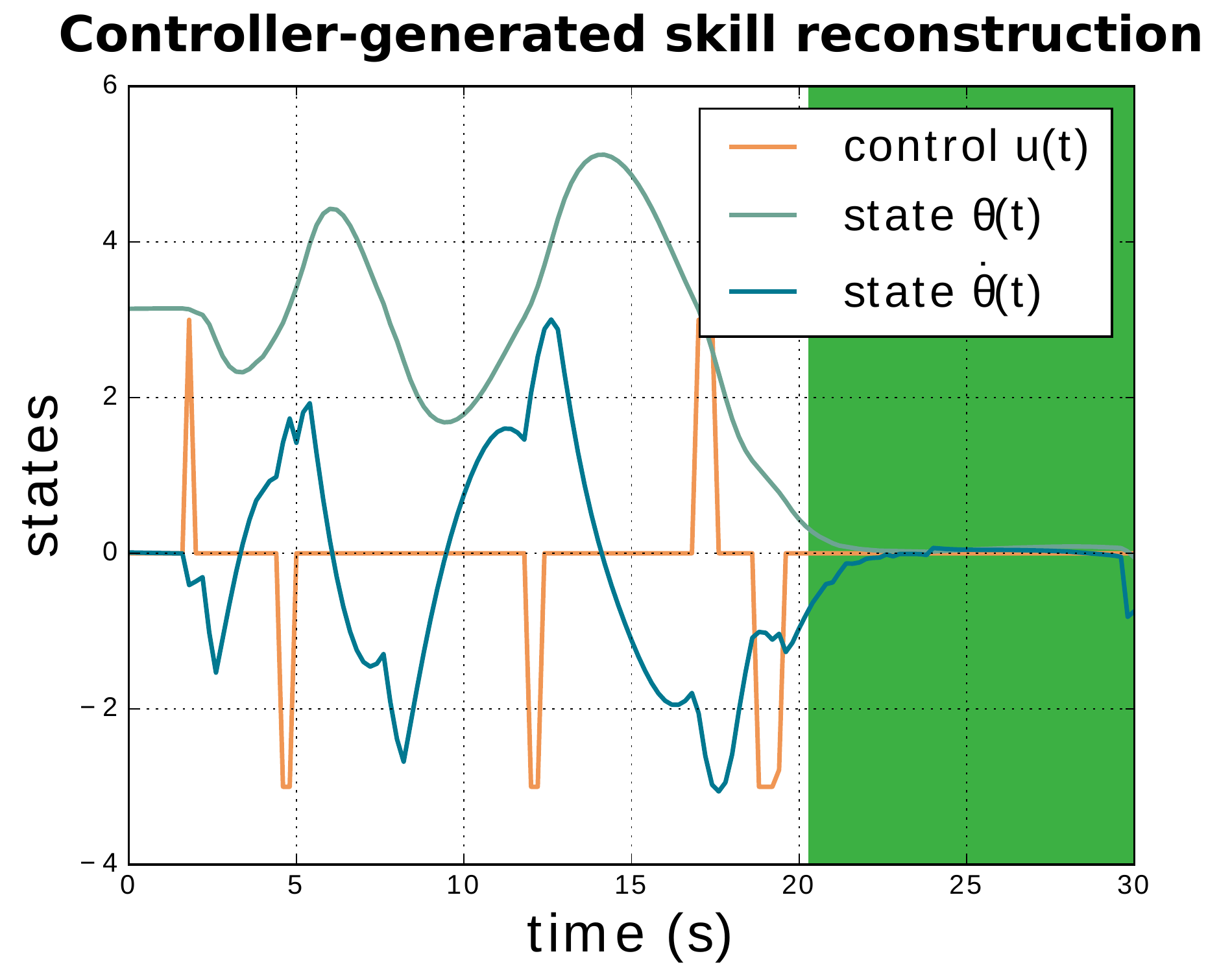}
    \end{subfigure} \hfill
    \begin{subfigure} \centering
    \includegraphics[width=0.3\textwidth]{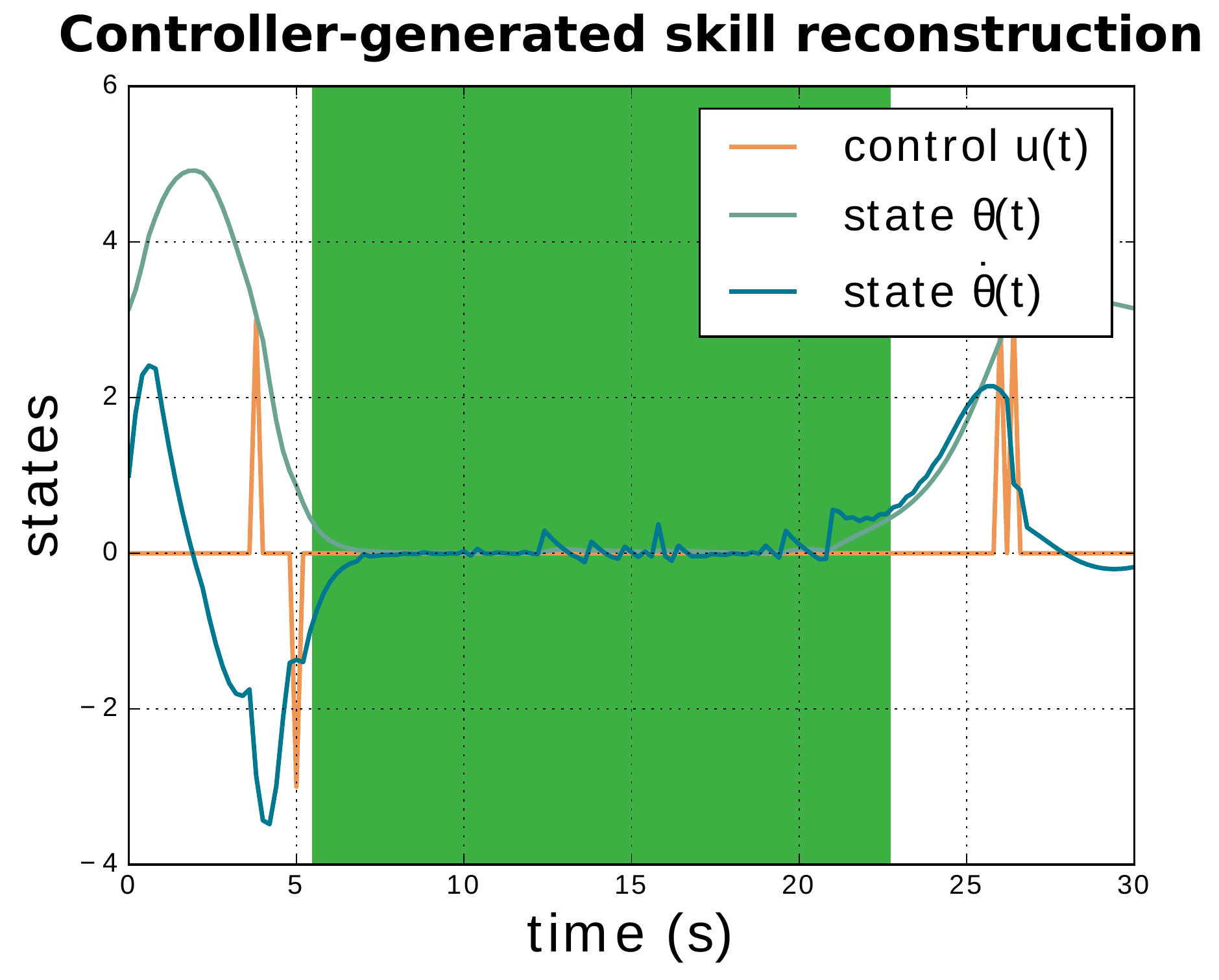} 
    \end{subfigure}%
    \caption{(left) Example task definition and skill reconstruction learned from 3 novice trajectories from Subject 6. Note that the controller just barely succeeds (for less than 5 seconds), exhibiting comparable performance to the original demonstrations, which had an average success time of 5.6 seconds. (center) Example task definition and skill reconstruction learned from 3 \textit{negative} demonstrations. Note that a negative demonstration includes only failed task attempts and---with this low-dimensional task---suffices for learning a sub-optimal, yet successful task definition. (right) Example task definition and skill reconstruction learned from positive and negative demonstrations from Subject 6. Note that while the \textit{posonly} controller exhibits performance similar to the original demonstrations, the \textit{posneg} controller significantly outperforms them. Although Subject 6 is still a novice at this task, we can learn a task representation comparable to the one learned from an expert trajectory (see Fig.~\ref{fig: pos-expert}) by soliciting both positive and negative demonstrations.}
  \label{fig: combined}
\end{figure*}

\section{Methods}
\label{sec:methods}

\subsection{Ergodic Task Definitions and Control}
In ergodic imitation, we generate a representation of an unknown task using spatial statistics. Since we avoid specifying temporal dependencies, we synthesize robotic controls that successfully achieve a task without necessarily replicating the demonstration's time-evolving trajectory. We define a single demonstration, represented as $d_i$, as the distribution of points in the state space making up the the state trajectory $x(t)$ for a given set of time $t \in [t_0, t_f ]$, and the set of demonstrations as $D = {d_1, ..., d_m}.$ This set may contain both positive and negative demonstrations, so we also store a label array $E = {e_1, ..., e_m}$ corresponding in length to $D$.

\textbf{Learning from positive demonstrations.}
A positive demonstration is defined to be a person's attempt at a task that is at least somewhat successful. Oftentimes user-provided positive demonstrations are incomplete or highly sub-optimal with multiple attempts at the task and corrective actions within the demonstration. This was the case in the user studies used in this work. 

During the task learning process, we use the demonstration trajectories to generate a task definition $\phi(x)$ by representing the spatial statistics of each demonstration trajectory $d$ with the Fourier decomposition $\phi_k$, as described in Eq.~\ref{ck}. We then average the $\phi_k$ values of the demonstrations to represent the collective spatial statistics of all the demonstrations. Regions of the state space where more time is spent in the trajectories have a higher density in the distribution than regions where less time is spent. Examples of the cart-pole inversion task learned from positive demonstrations---of an expert and a novice---are showed in Fig.~\ref{fig: pos-expert} and Fig.~\ref{fig: combined}, respectively.

To generate the distribution $\phi(x)$ from the demonstration trajectories $x(t)$, we calculate spatial Fourier coefficients of $x(t)$ using Fourier basis functions of the form
\begin{equation}
    F_k(x)=\frac{1}{h_k}\prod_{i=1}^n cos\left( \frac{k_i\pi}{L_i}x_i\right),
\end{equation}
where $k$ is a multi-index over $n$ dimensions, $h_k$ is a normalizing factor~\cite{mathew2011}, and $L_i$ is a measure of the length of the dimension. We then compute the coefficients of a time-averaged trajectory using Eq.~\ref{ck}.
\begin{equation}
    c_k = \frac{1}{T}\int_0^T F_k(x(t))dt\label{ck} 
\end{equation}
The coefficients of the demonstration trajectories are combined to form the coefficients that describe the task definition. We use a weighted average
\begin{equation}
    \phi_k = \sum_{j=1}^m w_jc_{k,j}\label{phik},
\end{equation}
where the weighting factor $w_j$ normalizes each demonstrations based on either the length of the trajectory or the relative quality of the demonstration. In the examples in this paper, we normalize by the length. Note that other representations of a distribution could be used instead of Fourier coefficients, including wavelets or Gaussian Mixtures, as long as a comparison metric of two distributions can be defined and meets the conditions for global convexity. 

\textbf{Learning from negative demonstrations.}
In this work, we employ negative demonstrations, defined as both unsuccessful task attempts and explicit demonstrations of what \textit{not} to do. Negative demonstrations can include good-faith attempts at a task where the demonstrator performs poorly, or explicit examples of actions that are far from the desired behavior. In the case of reaching a target with object avoidance, a negative demonstration might repeatedly circle the area of the object without reaching the target. Negative demonstrations add most complementary information to the positive (successful) demonstrations.

As with positive demonstrations, the demonstrated trajectories are represented by the Fourier decomposition $c_k$ calculated using Eq.~\ref{ck}. However, they are combined through subtraction---$w_j<0$ in Eq.~\ref{phik}---such that regions of the state space where more time is spent in the trajectories have a lower density in the distribution than regions where less time is spent. When only negative demonstrations (\textit{negonly}) are used, a uniform distribution is introduced into the demonstration set D and given a positive weight, $w_j>0$.
 
\begin{algorithm}
\caption{Ergodic Control Algorithm for LfD}
\begin{algorithmic}
\Require{ initial time $t_0$, initial state $x_0$, set of demonstrations $D =\{d_1,...,d_m\}$ with positive/negative labels  $\{e_1,...,e_m\}$, final time $t_f$}
\Ensure{ergodic trajectory $x(t) \rightarrow X$}
\State{\textbf{Define: } ergodic cost weight $Q$, highest order of coefficients $K$, control weight $R$, search domain bounds $\{L_1,...L_n\}$, sampling time $t_s$, time horizon $T$} 
\State{\textbf{Initialize: } nominal control $u_{nom}$, $i=0$}
\State{Generate distribution $D(s)$ from demonstration set $D$.}
\State{Calculate $\phi_k$ from distribution $D(s)$}
\While{$t_i < t_f$}
    \State{Compute $u_i^*$ using MPC}
    \State{Apply $u_i^*$ for $t \in [t_i, t_i+t_s]$ to get $x \forall t \in [t_i, t_i+t_s]$.}
    \State{Define $t_{i+1} = t_i+t_s, x_{i+1} = x(t_{i+1})$}
    \State{$i \leftarrow i+1$}
\EndWhile
\end{algorithmic}
\label{alg:eSAC}
\end{algorithm}

\textbf{Ergodic Control.} Once the task is learned---by combining demonstrations using Eq.~(\ref{phik})---, we use a model predictive controller (MPC) to synthesize controls that generate a trajectory to match the spatial statistics of the distribution representing the demonstration set. In defining the task objective, we use ergodicity, which relates the temporal behavior of a signal to a pre-defined distribution. Ergodicity can be measured by several metrics~\cite{scott2009,scott2013}; here we use the spectral approach~\cite{mathew2011}, which characterizes ergodicity by comparing spatial Fourier coefficients of $x(t)$ to coefficients of $\phi (x)$.
Assume we have an autonomous agent whose movements are governed by a dynamic model that is either known \textit{a priori} or learned from data and is of the form
\begin{equation}
    \dot{x} = f(x,u) = g(x) + h(x) u
\end{equation}
where $x\in \mathbb{R}^n$ is the state of the agent and $u \in \mathbb{R}^m$ is the control input or ``actions'' the robot can take. 
A trajectory $x(t)$ is \emph{ergodic} with respect to a distribution $\phi(x)$ if, for every neighborhood $\mathcal{N} \subset \mathcal{X}$, the amount of time $x(t)$ spends in $\mathcal{N}$ is proportional to the measure of $\mathcal{N}$ provided by $\phi (x)$. On a long enough time horizon, measuring a perfectly ergodic $x(t)$ gives a complete description of $\phi (x)$. Here, we ask that $x(t)$ be \emph{maximally ergodic}, by introducing a metric on the distance from ergodicity into the objective function, so that when $x(t)$ captures the statistics of $\phi (x)$ in a specified time horizon $T$, the metric is lower. Ergodicity can be quantified as the sum of the weighed square distance between Fourier coefficients of the distribution $\phi_k$ and the coefficients representing the trajectory $c_k$:
\begin{equation} \label{ergmet}
    \varepsilon = \sum_{k_1=0}^{K}...\sum_{k_n=0}^{K} \Lambda_k |c_k-\phi_k|^2,
\end{equation}
where there are $n$ dimensions and $K+1$ coefficients along each dimension, and the coefficients $c_k$ can be calculated using Eq.~\ref{ck}. The coefficient $\Lambda_k= \frac{1}{(1+||k||^2)^s}$, where $s=\frac{n+1}{2}$, places larger weights on lower frequency information.
 
We define the task objective as 
\begin{equation}
    J =  q \varepsilon + \int_0^T \frac{1}{2} u(t)Ru(t) dt
\end{equation}
with a cost to minimize the ergodic metric and a cost on the control effort used over time. 

Now, using the ergodic objective function defined above, we frame the control problem as an MPC problem, following work in~\cite{mavrommati2018real}. The full algorithm is outlined in Algorithm~\ref{alg:eSAC}.

\begin{figure*}[t!]
\vspace{1em}
  \centering
  \includegraphics[width=0.76\textwidth]{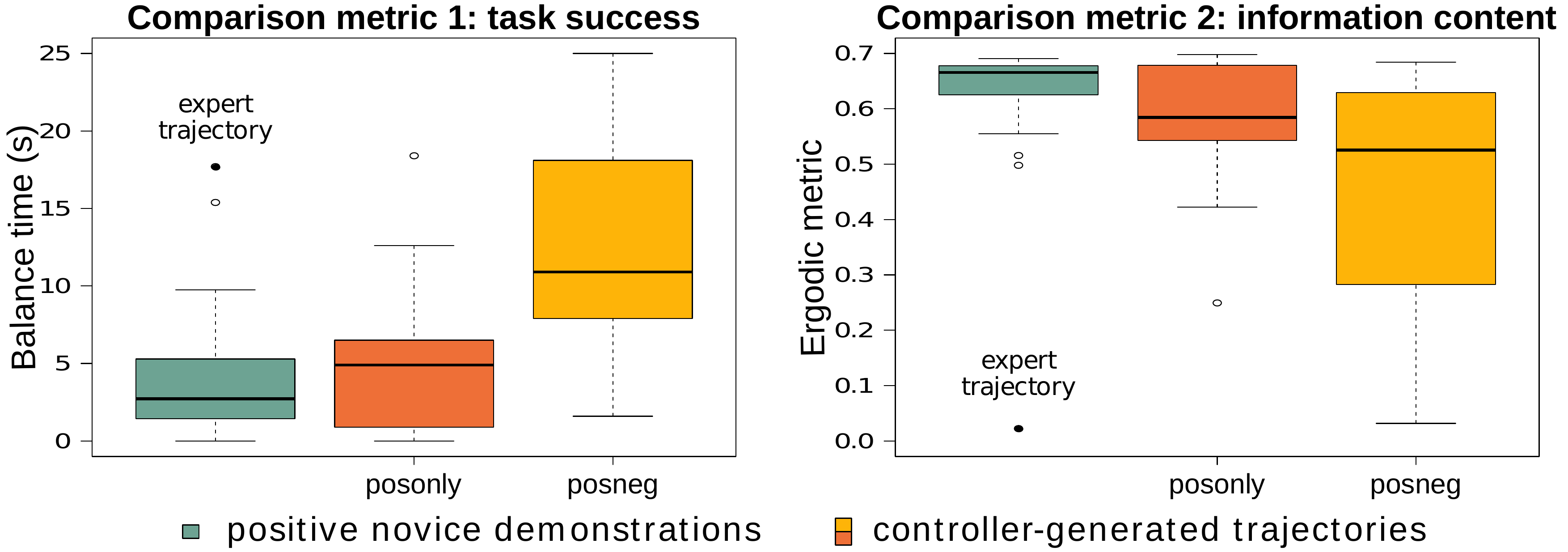}   
  \vspace{+3pt}
  \caption{Comparison of best task executions from 24 novice participants and skill reconstructions based on learned objectives, using only positive demonstrations (\textit{posonly}, orange) and using both positive and negative demonstrations (\textit{posneg}, yellow). We employ two performance metrics for the comparison: task success time (left) and the ergodic metric~\cite{ergodicitySR} (right), which measures information captured about the task in the learned distributions by comparing it to the true task definition. For both metrics, \textit{posonly} skill reconstructions achieve performance comparable to or better than the novice demonstrations. \textit{Posneg} trajectories significantly outperform the provided novice demonstrations in both metrics: 1 (F=9.07,~p=5.7e-10) and 2 (F=1.2,~p=3.8e-5)---in fact they provide skill reconstructions comparable to expert task executions. }
  \label{fig: pos-neg comparison}
\end{figure*}

\subsection{Experimental platforms}

We use two simulated experimental platforms and three benchmark tasks for algorithm validation. Similar to~\cite{torabi2018, brys2015, yang2019imitation}, we employ a cart-pole system for initial validation with end users. Inverting and balancing the cart-pole is a great example of a task that is reliably difficult for people, particularly novices, to accomplish. We also test two household tasks on a robot arm. These include reaching with object avoidance (similar to~\cite{palan2019learning, biyik2020learning}) as well as cleaning or wiping a surface (similar to~\cite{elliott2017learning, alizadeh2014learning}). These are good examples of real-world assistive tasks that encounter high variability in task execution during demonstrations. 

\textbf{Cart-pole System.} A simulated cart-pole with state vector $x=[\theta, \dot{\theta}, x_c, \dot{x_c}]$ and input $\ddot{x_c}$ was used in a previous study of 24 participants. Participants were each given 3 sets of 30 30-second attempts to invert the pole from its resting state to the unstable equilibrium. The data from this experiment---details of which can be found in~\cite{RSS2018-sharedcontrol} and~\cite{IJRR-HSC2020}---are used as the novice task demonstrations in this work.

For cart-pole inversion, a demonstration is defined as successful when during the 30-second attempt the participant reaches a state near the unstable equilibrium, specifically $|\theta|<0.4$~rad and $|\dot{\theta}|<0.75$~rad/s. We take the best demonstrations each user provided in set 3 (on average the best set) as positive demonstrations and take unsuccessful demonstrations from set 1 (on average the worst set) as negative demonstrations. Finally, we also test our approach on expert demonstrations---the positive expert demonstrations are generated using an optimal controller, whereas the negative expert demonstrations are generated by one of the authors. The true task definition for cart-pole inversion is defined as a Dirac delta function around $[\theta,\dot{\theta}]=[0,0].$

\textbf{Robot Arm Simulator.} We develop a pybullet simulation of the Franka Emika Panda Robot Arm to evaluate ergodic imitation on basic table-top tasks, specifically target reaching and table cleaning. In the simulation, we generate demonstrations for robot motion using a keyboard control interface. Keyboard keys control the desired end-effector position of the robot in the $[x, y]$ dimensions at a fixed end-effector height $z_d$. Demonstrations consist of the end-effector trajectories, from which we learn a task distribution. After that task definition is learned, we use ergodic MPC as a motion planner for the end-effector by generating desired end-effector trajectories $[x, y, z_d]$. For the ergodic controller, we model the system as a double-integrator with state $X = [x, y, \dot{x}, \dot{y}]$ and $U = [\ddot{x}, \ddot{y}]$. We use the an IK solver to generate joint states corresponding to the target trajectory and employ a low-level joint controller to execute the trajectory on the robot arm. 

For the target-reaching task, we define success as reaching a target location without colliding with an obstacle. For the cleaning task, success $m$ is evaluated as a continuous variable based on both workspace coverage and object avoidance. If the controller-generated trajectory comes too near the object, the trial is considered a failure ($m=0$). Otherwise, the cleaning is assessed by calculating the workspace visited by the end-effector, as approximated on a $5 \times 5$ grid.

\begin{figure*}
\vspace{0.5em}
    \centering
    \includegraphics[width =0.78\textwidth]{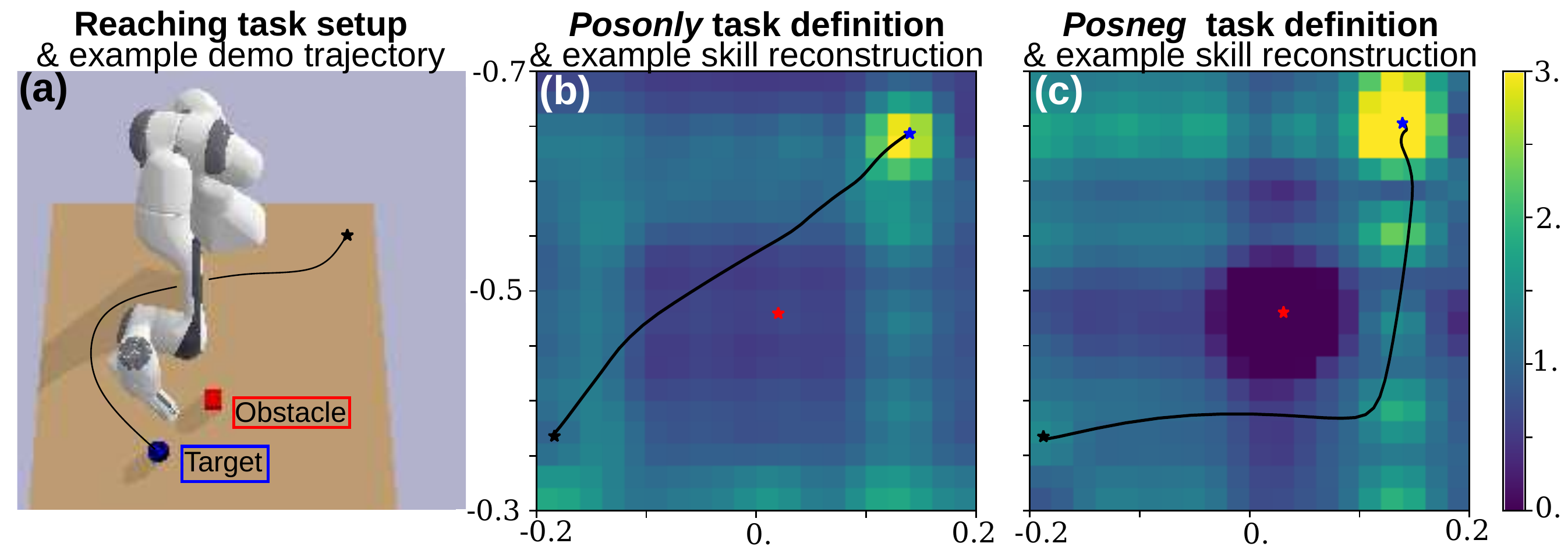}
    \caption{(a) Reaching towards a target (shown in blue) while avoiding an obstacle (shown in red) with a robot arm. Task definition and resulting robot end-effector trajectory generated with an ergodic controller using the positive-only demonstrations (b) and using combined positive and negative demonstrations (c). Negative demonstrations more effectively reflect the region of avoidance, representing what \textit{not} to do.}
    \label{fig:target-reacharound-result}
\end{figure*} 

\begin{figure*}
    \centering
    \includegraphics[width =0.9\textwidth]{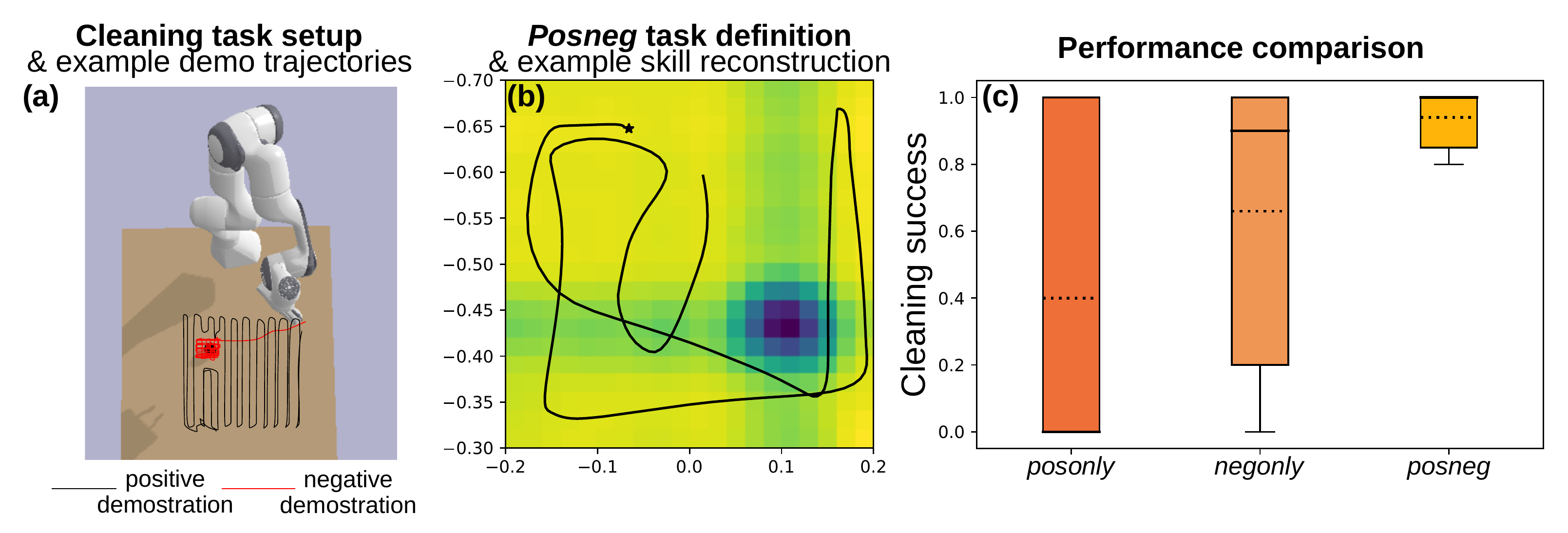}
    \caption{Cleaning around an object with a robot arm. (a) An example positive demonstration trajectory is shown in black and a negative demonstration is shown in red. (b) Task definition and resulting robot end-effector trajectory generated using the ergodic controller with positive and negative demonstrations. (c) Comparison of success between results from task definitions generated with positive, negative, and \textit{posneg} demonstrations. The black line represents the median result and the dotted line represents the mean. The \textit{posneg} definition results in significantly better performance than either \textit{posonly} or \textit{negonly} definitions, capturing both the desired cleaning and object avoidance goals. }
    \label{fig:obj-avoidance-result}
\end{figure*}

\section{Experimental Results}
\label{sec:result}

\subsection{Ergodic imitation enables learning from imperfect demos}

We show that ergodic imitation can be used to infer the cart-pole inversion task from imperfect novice demonstrations and that it can recreate the skill on average better than presented during demonstrations. More specifically, a t-test comparison shows that trajectories generated using ergodic imitation have higher success times (F=0.24,~p=0.06) and are more ergodic w.r.t. the true task definition (F=0.68,~p=0.002) than the provided demonstrations (see Fig.~\ref{fig: pos-neg comparison}). This means that when learning from only positive demonstrations, our trained controller will on average perform comparably or outperform the provided task demonstrations.

\subsection{Negative demos consistently improve learning}

Furthermore, we show that negative demonstrations add more value than numerous positive demonstrations, allowing data-efficient learning. For each of the 24 participants, we learn a task definition from 3 positive and 3 negative demonstrations. Again, we compare the controller-generated trajectories with the provided trajectories, using success time and the ergodic metric. Results of a t-test show that trajectories generated using ergodic imitation have higher success times (F=9.07,~p=5.7e-10) and are more ergodic with respect to the true task definition (F=1.2,~p=3.8e-5) than the provided demonstrations. They also have higher success times (F=1.4,~p=4.9e-7) and are more ergodic (F=0.79,~p=0.003) than the trajectories generated using \textit{posonly} demonstrations. Finally, note that the effect sizes are significantly larger than in the earlier comparison as visible in Fig.~\ref{fig: pos-neg comparison}.

In the event that an end-user cannot generate any successful demonstrations, we also demonstrate the ability to define a successful task specification from just negative demonstrations. This is visible in Fig.~\ref{fig: combined}, where we note that the skill reconstruction achieves inversion quite late around $t=20s$. Although it is likely impractical for most tasks to learn from only negative demonstrations, this interesting result illustrates that valuable information about a task can be conveyed through negative demonstrations. 

\subsection{Ergodic imitation with negative demos works for multimodal and multi-objective tasks}

Ergodic imitation with \textit{posneg} demonstrations extends to a variety of tasks. It is particularly useful for open-ended tasks, such as cleaning, where multiple successful task executions exist, or for multi-objective tasks, such as reaching a target while avoiding an object, where an added safety constraint is present. Negative demonstrations can be helpful, particularly when trying to represent constraints in the environment. 

We use ergodic imitation to learn to reach a target while avoiding an object. In Fig.~\ref{fig:target-reacharound-result}, we present an example trajectory generated from random initial conditions based on a task definition learned from 13 positive demonstrations and combined \textit{posneg} demonstrations (13 positive + 3 negative), all between 10-15 seconds in length. Note that with positive demonstrations, the goal location is successfully reflected, but the region of obstacle avoidance is only starting to appear. When we add negative demonstrations, the region of avoidance is more clearly defined.

We use ergodic imitation to clean a surface around an object. We compare the results for 10 controller-generated trajectories from random initial states for each type of task definition (\textit{posonly}, \textit{negonly}, and \textit{posneg})---generated from 5 positive, 2 negative, and 3+2 combined \textit{posneg} demonstrations, respectively. The positive demonstrations were 40-50 seconds in length whereas the negative ones---18-25 seconds. As visible in Fig.~\ref{fig:obj-avoidance-result}, the combination of positive and negative demonstrations offers best performance, highlighting the region to avoid while still representing the rest of the cleaning task. The \textit{posonly} controller-generated trajectories sometimes result in collisions. The \textit{negonly} skill reconstructions result in very few object collisions but the overall workspace coverage is low. The \textit{posneg} task definition significantly outperforms both other task specifications, resulting in no failures and a median 100\% success rate.

%==================================================================

\section{Conclusions \& Discussion}
\label{sec:conclusion}

This paper introduces ergodic imitation for learning from novice robot users and illustrates the value of negative demonstrations---reflecting what \textit{not} to do---in imitation learning. While positive-only demonstrations can result in successful skill reproduction, the combination of positive and negative demonstrations can help to efficiently generate task definitions for difficult tasks. Ergodic imitation is particularly well suited for multi-objective and open-ended tasks, where either multiple goals are equally important (e.g., moving a cup without spilling) or different motion trajectories can accomplish the same task. 

There is potential to extend the approach to applications with a focus on safety constraints and user preferences, such as assisted driving---similarly to~\cite{sadigh2017active} but without the need for preference querying. Moreover, in future work, the ergodic learning framework could be further automated by combining it with feature selection algorithms, such as~\cite{kroemer2017feature,pais2013learning,niekum2012iros}. Overall, the presented results are promising---the proposed algorithmic framework and negative demonstrations have potential to enable demonstration-efficient LfD from imperfect demonstrations for a range of robotic applications.

% \addtolength{\textheight}{-0cm}  

\bibliography{references} 
\bibliographystyle{plain}

\end{document}